\documentclass[a4paper]{article}

\usepackage{INTERSPEECH2019}
\usepackage{multirow}
\usepackage{changepage}

\title{Contextualizing ASR Lattice Rescoring with Hybrid Pointer Network Language Model}
\name{Da-Rong Liu$^{\dagger}$, Chunxi Liu$^\star$, Frank Zhang$^\star$, Gabriel Synnaeve$^\star$, Yatharth Saraf$^\star$,  Geoffrey Zweig$^\star$
\thanks{$\dagger$ Work was done when Da-Rong was an intern at Facebook.}  }

\address{
  $^\dagger$National Taiwan University  \quad  $^\star$Facebook AI, USA}
\email{   f07942148@ntu.edu.tw  \quad    \{chunxiliu,frankz,gab,ysaraf,gzweig\}@fb.com}

\begin{document}
\ninept

\maketitle
\begin{abstract}
Videos uploaded on social media are often accompanied with textual descriptions. In building automatic speech recognition (ASR) systems for videos, we can exploit the contextual information provided by such video metadata. In this paper, we explore ASR lattice rescoring by selectively attending to the video descriptions. 
We first use an attention based method to extract contextual vector representations of video metadata, and use these representations as part of the inputs to a neural language model during lattice rescoring. 
Secondly, we propose a hybrid pointer network approach to explicitly interpolate the word probabilities of the word occurrences in metadata. 
We perform experimental evaluations on both language modeling and ASR tasks, 
and demonstrate that both proposed methods provide performance improvements by selectively leveraging the video metadata. 
\end{abstract}

\section{Introduction}

Personalized or contextual automatic speech recognition, which aims to improve accuracy by leveraging additional information or external knowledge, has been an important research topic \cite{hall2015composition,mcgraw2016personalized,williams2018contextual,pundak2018deep,chen2019end}.
These prior works usually assume that a set of word-level biasing phrases are known ahead of time, e.g. a user’s personal contact list, which are used to nudge the ASR model towards outputting these particular phrases. 
In the conventional hybrid ASR framework, bias phrases can be compiled into a weighted finite state transducer (WFST), with vocabulary injection, and on-the-fly language model biasing techniques  \cite{hall2015composition, mcgraw2016personalized} have shown significant performance gains.
Similarly, for end-to-end ASR architectures like Listen, Attend and Spell (LAS)  \cite{chan2015listen}, the WFST representation of context $n$-grams is traversed along with the outputs from the LAS network, and the beam search decoding is biased either in first pass decoding \cite{chen2019end} or rescoring \cite{williams2018contextual}.
Alternatively, each context $n$-gram can also be embedded into a fixed dimensional representation, and such contextual information is summarized by an attention mechanism and further fed as an additional input to the decoder \cite{pundak2018deep, chen2019joint}.

\begin{figure}[t]
  \centering
  \includegraphics[width=7.0cm, scale=0.8]{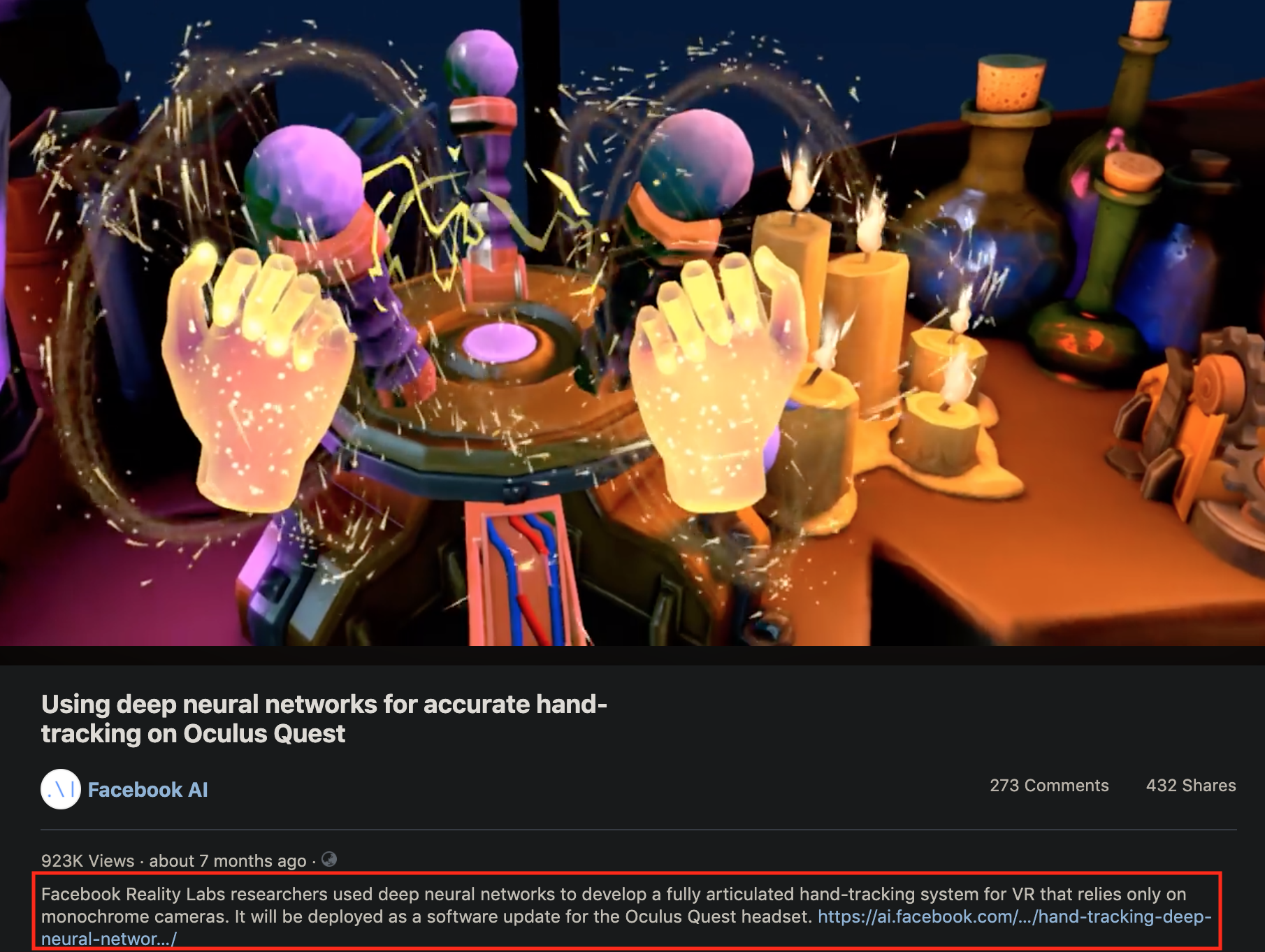}
  \caption{\textbf{Video metadata:} Social media videos are often associated with surrounding text, descriptions or titles, as denoted in the red box.
  }
  \label{fig:example}
\end{figure}

One main distinction in our work is that rather than using a list of named entities, such as user's contact lists or song names as in many prior approaches, here we aim to exploit contextual information from word sequences or paragraphs, as illustrated in Figure \ref{fig:example}. Henceforth, we refer to such textual content as \textit{video metadata}. Utilizing the video metadata effectively can be challenging,  since it not only contains potentially relevant information, but also irrelevant text.
To address this challenge, our neural LM training is explicitly conditioned on the video metadata, and selectively attends to the metadata via an attention mechanism. The resulting contextual neural LM is used to rescore the lattices generated from the first-pass hybrid ASR decoding.

Therefore, our proposed methods are similar to \cite{pundak2018deep} with three important distinctions. First, we produce the ASR lattice via a conventional WFST-based hybrid ASR model, and contextual biasing is performed by jointly rescoring the lattice and attending to the metadata. Second, rather than tying the contextual biasing with the end-to-end LAS training, we build the contextual LM separately from the acoustic model training, which effectively allows for modular evaluation and improvements of the contextual LM component. In common with earlier generations of technology, language model changes can be made independently of the acoustic model. Third, based on \cite{pundak2018deep} that utilized contextual information by an attention mechanism, we further propose the hybrid pointer network (which will be introduced in Section \ref{Hybrid Pointer Network}). 


While conventional sequence-to-sequence (seq2seq) models typically generate tokens from a predefined vocabulary \cite{bahdanau2014neural, nallapati2016abstractive}, 
pointer networks \cite{vinyals2015pointer} can be used to 
explicitly select and output tokens from the input (source) sequence.
Recently, hybrid pointer-generator networks (PGN), combining seq2seq models with pointer networks, have been proposed and used in summarization tasks \cite{see2017get}.
In such models, an additional scalar variable is generated at each time step and serves as a soft switch to choose between generating the token from a predefined vocabulary or selecting from the input sequence.
This has been shown to be particularly effective in generating rare words that have very few occurrences in training data \cite{see2017get}.

Our main contributions can be summarized into three categories:
\begin{itemize} 
\item 
We build a contextual language model that conditions on the video metadata.
We compare various alternatives for such a language model and demonstrate that a hybrid pointer network substantially outperforms all competing baselines in perplexities.
\item 
We then use this language model to rescore the lattice generated from the first pass ASR decoding.
We employed the pruned lattice rescoring algorithm \cite{xu2018pruned}, and show that after rescoring, our contextual LM performs better than all other baseline LMs in word error rate (WER).
\item 
We further perform analysis on how the quality of the video metadata affects the ASR performance.
\end{itemize}

\section{Contextual Language Model}


\begin{figure}[t]
  \centering
  \includegraphics[width=\linewidth]{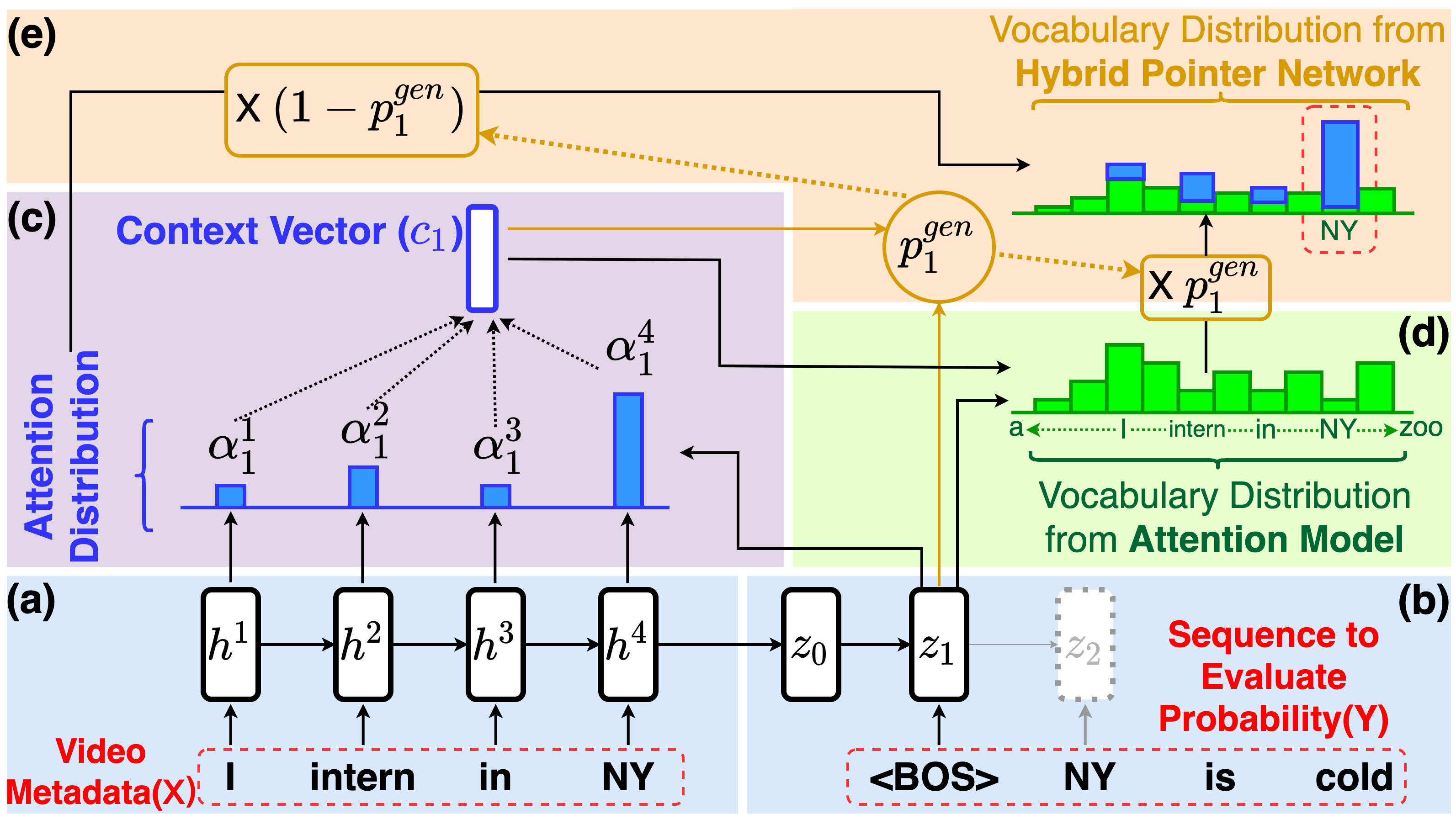}
  \caption{\textbf{Contextual Language Model Overview:} In this figure, video metadata is set to ``I intern in NY" and the sequence to evaluate probability is set to "NY is cold". At time-step one, the target of model is predicting P(``NY"). (a) The encoder, which encodes the video metadata into a sequence of hidden vectors $h^{i}$. (b) The decoder. (c) The attention mechanism, which generates the attention distribution over encoder hidden vectors, and summarizes them into a context vector. (d) The context vector and the decoder hidden state are used to generate the vocabulary distribution. (e) We can further interpolate the vocabulary distribution with attention distribution to boost the probability of rare word. In general, (a)+(b)+(c)+(d) form the attention model depicted in Section \ref{Attention Model}, and (a)+(b)+(c)+(d)+(e) form the hybrid pointer network depicted in Section \ref{Hybrid Pointer Network}.
}
  \label{fig:whole_model}
\end{figure}

A language model (LM) represents a probability distribution over sequences of $N$ tokens $\uppercase{y} = (  y_{1}, y_{2}, ...,y_{N} )$.
Given such a sequence, a LM assigns a probability to the sequence by modeling the probability of token $y_{k}$ conditioned on its history $\{  y_{1}, y_{2}, ...,y_{k-1} \}$.
The probability of the whole sequence can be decomposed as:
\begin{equation}
    P(y_{1}, y_{2}, ...,y_{N}) = \prod_{k=1}^{N}P(y_{k}|y_{1}, y_{2}, ...,y_{k-1})
\end{equation}
In this paper, our language model can be conditioning on the video metadata $\textbf{\uppercase{x}} = \{  x_{1}, x_{2}, ...,x_{M} \}$, where $M$ is the metadata sequence length.
Then, the conditional probability can be represented and decomposed as:
\begin{equation}
    P(y_{1}, y_{2}, ...,y_{N}|X) = \prod_{k=1}^{N}P(y_{k}|y_{1}, y_{2}, ...,y_{k-1},X)
\end{equation}
This is trained by minimizing the negative log-likelihood:
\begin{equation}
    \sum_{k=1}^{N}-log P(y_{k}|y_{1}, y_{2}, ...,y_{k-1},X;\theta_{model})
\end{equation}
where $\theta_{model}$ is the set of trainable parameters.
For modeling $P(y_{k}|y_{1}, y_{2}, ...,y_{k-1},X)$,
we first describe the attention model in Section \ref{Attention Model}, and
then we will show how to adapt the attention model to the hybrid pointer network in Section \ref{Hybrid Pointer Network}.

\subsection{Attention Model}\label{Attention Model}
The attention model, which is depicted as a component of the hybrid pointer network in Figure \ref{fig:whole_model}, is similar to the attention model proposed in \cite{luong2015effective} with only one difference.
The model proposed in \cite{luong2015effective} is targeted to the translation task, which will auto-regressively generate words at the decoder.
The auto-regressive generation means we will generate one word at one time-step, and feed the word as input to the next time-step.
However, as our model is a language model, the input to the decoder is the word sequence of which we evaluate the probabilities.
In this work, these word sequences will be the ASR hypotheses from the lattice.
We describe the detail formulation below.

The tokens of the metadata $x_{i}$ are fed one-by-one into the encoder (multi-layer LSTM), producing a sequence of hidden states $h^{i}$.
At each time step $t$, the inputs to the decoder are the word embedding of $y_{t}$ and the previous hidden state $z_{t-1}$, generating the current hidden state $z_{t}$. 
The attention distribution is computed as:
\begin{align}
 & \hat{\alpha}_{t}^{i} = (W_{z}z_{t}+b_{z})^{T}h^{i}
 \\
 & \alpha_{t} = softmax(\hat{\alpha}_{t})
\end{align}
Next, the attention distribution is used to produce a weighted sum of the encoder hidden states, known as the context vector:
\begin{equation}
    c_{t} = \sum_{i}\alpha_{t}^{i}h^{i}
\end{equation}
The context vector can be viewed as the summary of the encoder information, which is then concatenated with decoder hidden state $z_{t}$ and passed through two linear layers to determine the probability of the next word $y_{t+1}$:
\begin{align}
 & P_{vocab}(w) = softmax(W^\prime(W[z_{t};c_{t}]+b)+b^\prime) \label{eqn:vocab}
 \\
 & P(y_{t+1}|y_{1}, y_{2}, ...,y_{t},X) \notag \\
 & = P(y_{t+1}|z_{t},c_{t}) = P_{vocab}(y_{t+1}) \label{eqn:condition}
\end{align}
where $W$, $W^{\prime}$, $b$, $b^{\prime}$ are learnable parameters. 
$P_{vocab}$ is the distribution over the whole vocabulary.

\subsection{Hybrid pointer network}\label{Hybrid Pointer Network}
In the attention model, we have incorporated video metadata information while predicting the probabilities.
However, it will be difficult for the model to predict words of rare occurrences in the training data.
Take Figure \ref{fig:whole_model} as an example.
We assume the word ``NY" is rare in training data and the attention model is perfect in predicting $p(NY)$, which means:
\begin{align}
 & \alpha_{1}^{4} = 1 \notag \\
 & \alpha_{1}^{1} =  \alpha_{1}^{2} =  \alpha_{1}^{3} = 0 \notag 
\end{align}
In this way, we can get $c_{1}=h^{4}$.
Because ``NY" is rare in training data, there will be two consequences.
First, the word embedding of ``NY", i.e. $h^4$, may not be well trained.
Second, even if $h^4$ does contain the information of the word ``NY", according to equation (\ref{eqn:condition}):
\begin{equation}
    P(NY) = P(NY|z_{1}, c_{1}) = P(NY|z_{1}, h^{4})
\end{equation}
it can be difficult for the model to map $h^4$ back to ``NY", because again it is rare in training data.

To address this problem, we propose to use a hybrid pointer network similar to \cite{see2017get}.
In a hybrid pointer network, another random variable $P^{gen}_{t} \in (0,1)$ is introduced:
\begin{equation}
    P^{gen}_{t} = \sigma(W^{gen}[c_{t};z_{t};y_{t}]+b)
\end{equation}
This $P^{gen}_{t}$ is used as a soft switch to select between the distribution $P_{vocab}$ generated in equation (\ref{eqn:vocab}) or sample from the attention distribution $\hat{\alpha}_{t}$:
\begin{equation}
    P_{vocab}^{ptr}(w) = p^{gen}_{t}P_{vocab}(w) + (1-p^{gen}_{t})\sum_{i:w_{i}=w}\alpha_{t}^{i}
\end{equation}
When $p^{gen}_{t}$ is close to one, it means the language model has high confidence to directly generate the distribution.
In contrast, if $p^{gen}_{t}$ is close to zero, it means the model has low confidence, and turns to utilizing the information from metadata.
Finally, the probability of  the next word $y_{t+1}$ will be:
\begin{align}
P(y_{t+1}|y_{1}, y_{2}, ...,y_{t},X) = P_{vocab}^{ptr}(y_{t+1})
\end{align}

\section{Evaluation}
We evaluate the effectiveness of our proposed approaches on our in-house English (EN) and Spanish (ES) video datasets, which are sampled from public social media videos and de-identified before transcription. 
These videos contain a diverse range of speakers, accents, topics, and
acoustic conditions making automatic recognition difficult.
Each data instance consists of the audio, reference transcription and the corresponding video metadata text.
The test sets for each language are composed of \texttt{clean}, \texttt{noisy} and \texttt{extreme} categories, with 
\texttt{extreme} being more acoustically challenging than \texttt{clean} and \texttt{noisy}. 
The dataset sizes are shown in Table \ref{tab:dataset}.
\begin{table}[t]
\caption{Dataset sizes in hours}
\label{tab:dataset}
\begin{tabular}{ c c c c c c }
\hline
         &  Train  & Valid &  \multicolumn{3}{   c  }{Test }  \\ 
         &         &       &  clean   &  noisy &   extreme     \\
\hline \hline
English  & 14k (hrs)   &  9.7  &  20.2   &  18.6   & 49.1  \\ \cline{4-6} 
\hline \hline
Spanish  &  7.5k  & 9.9  &  17.2  &  19.5     &  46.1  \\ \cline{4-6} 
\hline \hline
\end{tabular}
\end{table}

We first evaluate the language model perplexities in Section \ref{exp:Language Modeling}.
Five different language models are compared:
\begin{itemize}
	\item 5-gram: the language model with Kneser-Ney smoothing and used in the first-pass ASR decoding.
	\item LSTM: a multi-layer LSTM.
	\item cache-LSTM: this is a simple way for the LSTM LM to leverage video metadata. 
	We interpolate the output distribution of LSTM LM with the unigram probability distribution of video metadata. 
	The interpolation weight is a tunable hyperparameter.
	\item attention model: the model described in Section \ref{Attention Model}
	\item hybrid pointer network: the model described in Section \ref{Hybrid Pointer Network}
\end{itemize}
Specifically, the $5$-gram and LSTM LMs do not use the information from the video metadata, while the remaining do.
A two-layer LSTM with 512 hidden units and 0.1 dropout are used as the recurrent part of all recurrent models.
Adaptive softmax \cite{grave2017efficient} is used for efficient training.
We implement all our recurrent LMs based on the fairseq toolkit \cite{ott2019fairseq}. Kaldi decoder \cite{Povey_ASRU2011} is used to produce ASR lattices. 
All language models, except the $5$-gram model, are then used to perform lattice rescoring. The WER comparisons are shown in Section \ref{exp:WER}.

\subsection{Language Modeling}\label{exp:Language Modeling}
We first evaluate the effectiveness of each LM in terms of perplexity.
LMs are trained using both transcriptions and video metadata from the \texttt{train} set.
We build our vocabulary as all words seen in the training data, while leaving remaining words as OOVs.
We use cosine learning rate (LR) scheduler \cite{loshchilov2016sgdr} and NAG optimizer with initial LR  0.001.
The results are shown in Table \ref{tab:ppl}.
\setlength{\tabcolsep}{0.17cm}
\begin{table}[]
\caption{Perplexities of each language model on the test sets. The values in the parenthesis for (c) and (d) denote the interpolation weights.}
\label{tab:ppl}
\begin{tabular}{|l|c|c|c|}
\hline
                    & \multicolumn{3}{c|}{English} \\ \hline
                    & \texttt{clean}     &  \texttt{noisy}    & \texttt{extreme}    \\ \hline
(a) 5-gram           & 129.9   & 150.1   & 150.4  \\ \hline
(b) LSTM             & 109.6   & 114.9  & 119.6  \\ \hline
(c) cache-LSTM (0.1) & 105.1   & 115.35  & 119.8  \\ \hline
(d) cache-LSTM (0.2) & 113.0   & 125.7  & 130.8  \\ \hline
(e) attention model   & 99.1     & 106.2  & 110.2  \\ \hline
(f) hybrid pointer network  & \textbf{76.9}    & \textbf{91.0}   & \textbf{95.2}  \\ \hline
                    & \multicolumn{3}{c|}{Spanish} \\ \hline
                    & \texttt{clean}     &  \texttt{noisy}    & \texttt{extreme}  \\ \hline
(a) 5-gram           & 176.4   & 194.0  & 209.6  \\ \hline
(b) LSTM             & 119.0   & 130.3  & 151.6  \\ \hline
(c) cache-LSTM (0.1) & 118.0      & 137.2  & 160.4  \\ \hline
(d) cache-LSTM (0.2) & 127.6   & 151.7  & 177.8  \\ \hline
(e) attention model   & 107.4   & 118.9  & 139.7  \\ \hline
(f) hybrid pointer network  & \textbf{84.4}    & \textbf{101.4}  & \textbf{121.2}  \\ \hline
\end{tabular}
\end{table}

We can see there are  significant improvements from the $5$-gram LM to each of the recurrent neural LMs, i.e., (a) vs (b), (c), (d), (e), and (f).
Comparing the cache-LSTM with the LSTM ((b) vs (c), (d)), we see that even though the cache-LSTM does leverage the video metadata, it does not perform better than the LSTM trained only on transcripts. This indicates that naively interpolating with the unigram distribution of the metadata may not be helpful.
The attention model performs better than LSTM and cache-LSTM ((e) vs (b), (c), (d)), because in attention LSTM, the model automatically learns how to leverage the video metadata.
Finally, the hybrid pointer network performs best, as it can overcome the shortcomings of the attention model as described in Section \ref{Hybrid Pointer Network}.

\begin{figure*}[t]
  \centering
  \includegraphics[width=15.1cm, scale=0.8]{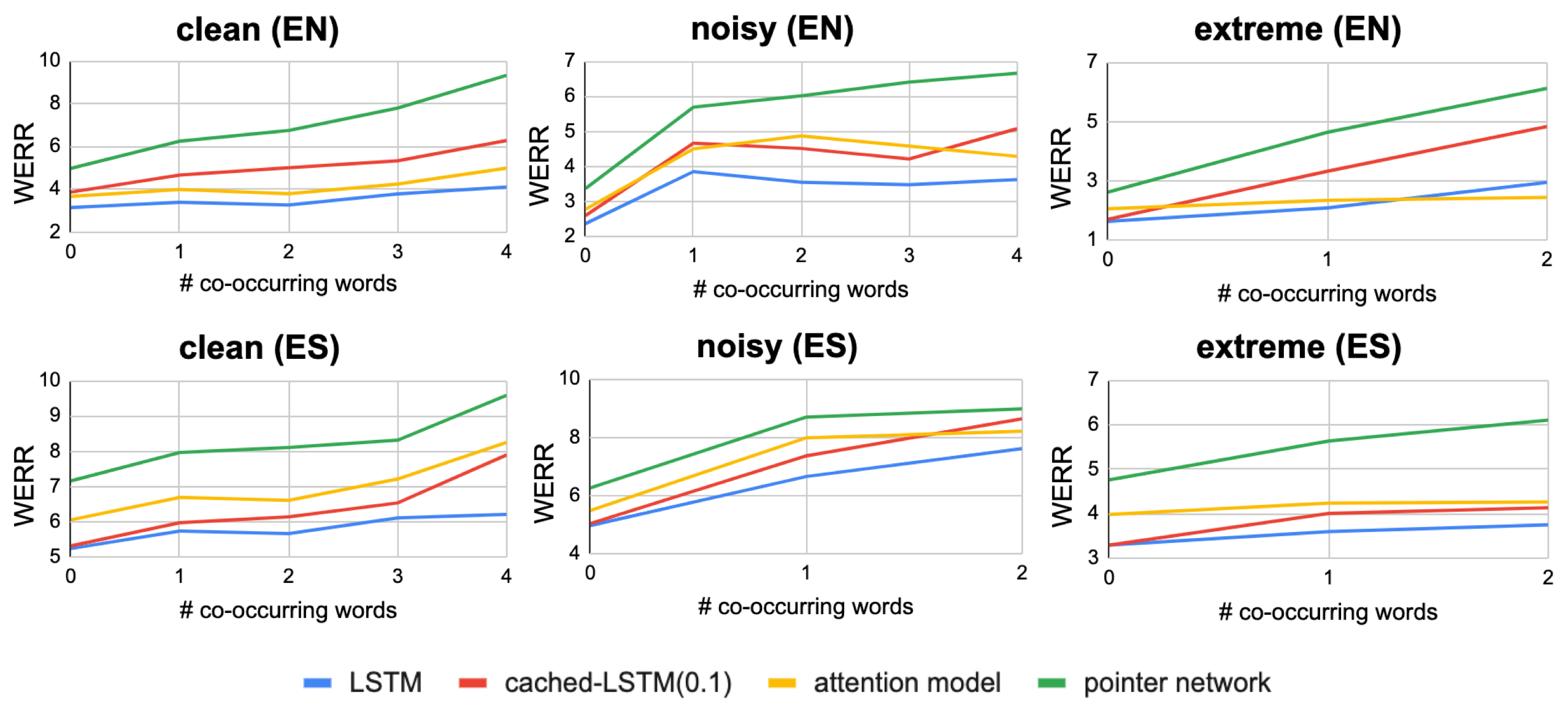}
  \vspace{-0.2cm}
  \caption{Analysis: in this figure, we report the WERR as the number of co-occurring words between video metadata and transcription varies.
  WERR denotes the relative WER reduction compared to the WER of first-pass decoding.
  We only report the number if the remaining test data size is more than half an hour, 
  since there could be too much variance if the remaining test data size is too small.
  }
  \label{fig:analysis}
\end{figure*}

\subsection{ASR performance}\label{exp:WER}
We now evaluate the effectiveness of the proposed method on the ASR task.
We first produce lattices via first-pass decoding with a graphemic hybrid ASR system \cite{le2019senones}.
For acoustic modeling, we utilize a hybrid ASR model with a graphemic lexicon trained with the lattice-free MMI criterion \cite{povey2016purely}.
In the first-pass decoding, we use Kaldi decoder with 5-gram LM (from Table\ref{tab:ppl} (a)) to generate lattices, with lattice beam 8.
The lattices are then rescored with the neural LMs by the pruned lattice algorithm \cite{xu2018pruned}. 
A 5-gram approximation is adopted \cite{xu2018pruned, liu2016two} to reduce the search space, i.e., we merge search paths containing the same last $5$ history words.

The results are shown in Table \ref{tab:wer}.
As expected, neural lattice rescoring improves the first-pass decoding results ((a) vs (b)).
The cache-LSTM does marginally better than LSTM ((b) vs (c)), but the improvement is small and unstable.
While attention model can further boost the performance, the hybrid pointer network achieves the best performance.

\setlength{\tabcolsep}{0.16cm}
\begin{table}[!htb]
\caption{ASR results in WER.
In rescoring with neural LMs, we interpolate the neural LM and $n$-gram LM scores as in \cite{xu2018pruned}.
The weight is tuned on \texttt{valid} set, and is shown in parenthesis for each language.}
\label{tab:wer}
\begin{tabular}{|l|c|c|c|}
\hline
                 & \multicolumn{3}{c|}{English (0.6)} \\ \hline
                 & \texttt{clean}     &  \texttt{noisy}    & \texttt{extreme}   \\ \hline
(a) first-pass       & 15.3   & 22.0   & 28.3    \\ \hline
(b) LSTM             & 14.81   & 21.47   & 27.8   \\ \hline
(c) cache-LSTM (0.1) & 14.7    & 21.4   & 27.8   \\ \hline
(d) attention model  & 14.7   & 21.4   & 27.7   \\ \hline
(e) hybrid pointer network     & \textbf{14.5}   & \textbf{21.3}   & \textbf{27.6}   \\ \hline
                 & \multicolumn{3}{c|}{Spanish (0.7)} \\ \hline
                & \texttt{clean}     &  \texttt{noisy}    & \texttt{extreme}   \\ \hline
(a) first-pass       & 13.6   & 15.5    & 21.9   \\ \hline
(b) LSTM             & 12.8   & 14.7   & 21.1   \\ \hline
(c) cache-LSTM (0.1) & 12.8   & 14.7   & 21.1   \\ \hline
(d) attention model  & 12.7   & 14.7   & 21.0   \\ \hline
(e) hybrid pointer network    & \textbf{12.6}   & \textbf{14.5}   & \textbf{20.8}   \\ \hline
\end{tabular}
\end{table}

\setlength{\tabcolsep}{0.27cm}
\begin{table}[!htb]
\begin{center}
\caption{
The proportion of test data that has none of video metadata in each test category.}
\vspace{-0.2cm}
\label{tab:empty}
\begin{tabular}{ c c c c }
\hline
 & \texttt{clean}    & \texttt{noisy}    & \texttt{extreme}    \\ 
\hline 
English & 78 / 1203 & 296 / 903 & 3214 / 7092 \\ \hline \hline
Spanish & 53 / 732  & 452 / 821 & 971 / 2026  \\ \hline \hline
\end{tabular}
\end{center}
\end{table}
\vspace{-0.2cm}

\subsection{Analysis}
Although we have shown the feasibility of hybrid pointer network in Section \ref{exp:WER},
we can see that the relative WER reduction (WERR) from LSTM to pointer LSTM is stable, but up to 2\% (Table \ref{tab:wer} (b) vs (e)).
The improvements can be correlated with the video metadata quality.
Table \ref{tab:empty} shows there is a large portion of dataset where the video metadata is absent. 
While the video metadata text could be irrelevant to the video transcriptions,
being  null  will certainly limit its effectiveness. 
To understand the importance of the available metadata size, Figure \ref{fig:analysis} shows the results as the number of co-occurring words in both metadata and reference transcription varies.
In our analysis, we select the test data instances with 1, 2, 3 or 4 co-occurring words between transcription and video metadata.
The 3000 most frequent words are not counted as co-occurring words, because hypothetically a co-occurring high frequency word, like `the', may not indicate the video metadata quality.

The results are shown in Figure \ref{fig:analysis}.
As the number of co-occurring words increases, we first observe that,
the WERR given by hybrid pointer network is more significant than the gain seen in the overall test set.
Although cache-LSTM and attention-LSTM also utilize the video metadata information, the WERR is not as substantial as in the pointer network.
Also, comparing the pointer network and the LSTM models, the gap between the WERR curves increase in most cases as   the number of co-occurring words grows, 
which indicates the effectiveness of our proposed method given a reasonable video metadata quality.

\section{Conclusions}
In this work, we propose the use of a hybrid pointer network LM for lattice rescoring, thus making use of text metadata accompanying social media videos.
We analyze the conditions of its effectiveness, and demonstrate that it can provide improvements in both LM perplexity and ASR WER.
Also, in the hybrid pointer network framework, we can replace the recurrent components of multi-layer LSTM with other neural models, such as neural transformers \cite{irie2019language}.



\begin{thebibliography}{10}
\providecommand{\url}[1]{#1}
\csname url@samestyle\endcsname
\providecommand{\newblock}{\relax}
\providecommand{\bibinfo}[2]{#2}
\providecommand{\BIBentrySTDinterwordspacing}{\spaceskip=0pt\relax}
\providecommand{\BIBentryALTinterwordstretchfactor}{4}
\providecommand{\BIBentryALTinterwordspacing}{\spaceskip=\fontdimen2\font plus
\BIBentryALTinterwordstretchfactor\fontdimen3\font minus
  \fontdimen4\font\relax}
\providecommand{\BIBforeignlanguage}[2]{{%
\expandafter\ifx\csname l@#1\endcsname\relax
\typeout{** WARNING: IEEEtran.bst: No hyphenation pattern has been}%
\typeout{** loaded for the language `#1'. Using the pattern for}%
\typeout{** the default language instead.}%
\else
\language=\csname l@#1\endcsname
\fi
#2}}
\providecommand{\BIBdecl}{\relax}
\BIBdecl

\bibitem{hall2015composition}
K.~Hall, E.~Cho, C.~Allauzen, F.~Beaufays, N.~Coccaro, K.~Nakajima, M.~Riley,
  B.~Roark, D.~Rybach, and L.~Zhang, ``Composition-based on-the-fly rescoring
  for salient n-gram biasing,'' 2015.

\bibitem{mcgraw2016personalized}
I.~McGraw, R.~Prabhavalkar, R.~Alvarez, M.~G. Arenas, K.~Rao, D.~Rybach,
  O.~Alsharif, H.~Sak, A.~Gruenstein, F.~Beaufays \emph{et~al.}, ``Personalized
  speech recognition on mobile devices,'' in \emph{Proc. ICASSP}, 2016.

\bibitem{williams2018contextual}
I.~Williams, A.~Kannan, P.~S. Aleksic, D.~Rybach, and T.~N. Sainath,
  ``Contextual speech recognition in end-to-end neural network systems using
  beam search.''

\bibitem{pundak2018deep}
G.~Pundak, T.~N. Sainath, R.~Prabhavalkar, A.~Kannan, and D.~Zhao, ``Deep
  context: end-to-end contextual speech recognition,'' in \emph{2018 IEEE
  Spoken Language Technology Workshop (SLT)}.\hskip 1em plus 0.5em minus
  0.4em\relax IEEE, 2018, pp. 418--425.

\bibitem{chen2019end}
Z.~Chen, M.~Jain, Y.~Wang, M.~L. Seltzer, and C.~Fuegen, ``End-to-end
  contextual speech recognition using class language models and a token passing
  decoder,'' in \emph{Proc. ICASSP}, 2019.

\bibitem{chan2015listen}
W.~Chan, N.~Jaitly, Q.~V. Le, and O.~Vinyals, ``Listen, attend and spell,''
  \emph{arXiv preprint arXiv:1508.01211}, 2015.

\bibitem{chen2019joint}
Z.~Chen, M.~Jain, Y.~Wang, M.~L. Seltzer, and C.~Fuegen, ``Joint grapheme and
  phoneme embeddings for contextual end-to-end {ASR},'' in \emph{Proc.
  Interspeech 2019}, 2019.

\bibitem{bahdanau2014neural}
D.~Bahdanau, K.~Cho, and Y.~Bengio, ``Neural machine translation by jointly
  learning to align and translate,'' \emph{arXiv preprint arXiv:1409.0473},
  2014.

\bibitem{nallapati2016abstractive}
R.~Nallapati, B.~Zhou, C.~Gulcehre, B.~Xiang \emph{et~al.}, ``Abstractive text
  summarization using sequence-to-sequence rnns and beyond,'' \emph{arXiv
  preprint arXiv:1602.06023}, 2016.

\bibitem{vinyals2015pointer}
O.~Vinyals, M.~Fortunato, and N.~Jaitly, ``Pointer networks,'' in
  \emph{Advances in Neural Information Processing Systems}, 2015, pp.
  2692--2700.

\bibitem{see2017get}
A.~See, P.~J. Liu, and C.~D. Manning, ``Get to the point: Summarization with
  pointer-generator networks,'' \emph{arXiv preprint arXiv:1704.04368}, 2017.

\bibitem{xu2018pruned}
H.~Xu, T.~Chen, D.~Gao, Y.~Wang, K.~Li, N.~Goel, Y.~Carmiel, D.~Povey, and
  S.~Khudanpur, ``A pruned rnnlm lattice-rescoring algorithm for automatic
  speech recognition,'' in \emph{2018 IEEE International Conference on
  Acoustics, Speech and Signal Processing (ICASSP)}.\hskip 1em plus 0.5em minus
  0.4em\relax IEEE, 2018, pp. 5929--5933.

\bibitem{luong2015effective}
M.-T. Luong, H.~Pham, and C.~D. Manning, ``Effective approaches to
  attention-based neural machine translation,'' \emph{arXiv preprint
  arXiv:1508.04025}, 2015.

\bibitem{grave2017efficient}
E.~Grave, A.~Joulin, M.~Ciss{\'e}, H.~J{\'e}gou \emph{et~al.}, ``Efficient
  softmax approximation for gpus,'' in \emph{Proceedings of the 34th
  International Conference on Machine Learning-Volume 70}.\hskip 1em plus 0.5em
  minus 0.4em\relax JMLR. org, 2017, pp. 1302--1310.

\bibitem{ott2019fairseq}
M.~Ott, S.~Edunov, A.~Baevski, A.~Fan, S.~Gross, N.~Ng, D.~Grangier, and
  M.~Auli, ``fairseq: A fast, extensible toolkit for sequence modeling,'' in
  \emph{Proceedings of NAACL-HLT 2019: Demonstrations}, 2019.

\bibitem{Povey_ASRU2011}
D.~Povey, A.~Ghoshal, G.~Boulianne, L.~Burget, O.~Glembek, N.~Goel,
  M.~Hannemann, P.~Motlicek, Y.~Qian, P.~Schwarz, J.~Silovsky, G.~Stemmer, and
  K.~Vesely, ``The kaldi speech recognition toolkit,'' in \emph{IEEE 2011
  Workshop on Automatic Speech Recognition and Understanding}.\hskip 1em plus
  0.5em minus 0.4em\relax IEEE Signal Processing Society, Dec. 2011, iEEE
  Catalog No.: CFP11SRW-USB.

\bibitem{loshchilov2016sgdr}
I.~Loshchilov and F.~Hutter, ``Sgdr: Stochastic gradient descent with warm
  restarts,'' \emph{arXiv preprint arXiv:1608.03983}, 2016.

\bibitem{le2019senones}
D.~Le, X.~Zhang, W.~Zheng, C.~F{\"u}gen, G.~Zweig, and M.~L. Seltzer, ``From
  senones to chenones: Tied context-dependent graphemes for hybrid speech
  recognition,'' \emph{Proc. ASRU}, 2019.

\bibitem{povey2016purely}
D.~Povey, V.~Peddinti, D.~Galvez, P.~Ghahremani, V.~Manohar, X.~Na, Y.~Wang,
  and S.~Khudanpur, ``Purely sequence-trained neural networks for asr based on
  lattice-free mmi.'' in \emph{Proc. Interspeech}, 2016.

\bibitem{liu2016two}
X.~Liu, X.~Chen, Y.~Wang, M.~J. Gales, and P.~C. Woodland, ``Two efficient
  lattice rescoring methods using recurrent neural network language models,''
  \emph{IEEE/ACM Transactions on Audio, Speech, and Language Processing},
  vol.~24, no.~8, pp. 1438--1449, 2016.

\bibitem{irie2019language}
K.~Irie, A.~Zeyer, R.~Schl{\"u}ter, and H.~Ney, ``Language modeling with deep
  transformers,'' \emph{Proc. Interspeech}, 2019.

\end{thebibliography}
\end{document}